\documentclass[sigconf, 9pt]{acmart}

\renewcommand\footnotetextcopyrightpermission[1]{} % removes footnote with conference information in first column
\usepackage[font = small]{caption}
\usepackage{booktabs} % For formal tables
\usepackage{xcolor}
\usepackage[linesnumbered,ruled,vlined]{algorithm2e}

\SetCommentSty{mycommfont}

\makeatletter
\def\bstctlcite{\@ifnextchar[{\@bstctlcite}{\@bstctlcite[@auxout]}}
\def\@bstctlcite[#1]#2{\@bsphack
 \@for\@citeb:=#2\do{%
   \edef\@citeb{\expandafter\@firstofone\@citeb}%
   \if@filesw\immediate\write\csname #1\endcsname{\string\citation{\@citeb}}\fi}%
 \@esphack}
\makeatother

\vspace{-16pt}
\begin{document}

% \bstctlcite{IEEEexample:BSTcontrol}

%\title[shot title]{long title}
\title{DeepN-JPEG: A Deep Neural Network Favorable JPEG-based Image Compression Framework}
\author{
Zihao Liu$^1$, 
Tao Liu$^1$, 
Wujie Wen$^1$, 
Lei Jiang$^2$, 
Jie Xu$^3$,
Yanzhi Wang$^4$,
Gang Quan$^1$ \\
$^1$ Flordia International University,
$^2$ Indiana University, \\
$^3$ University of Miami, 
$^4$ Syracuse University \\
zliu021@fiu.edu, 
tliu023@fiu.edu, 
wwen@fiu.edu, \\
jiang60@iu.edu,
jiexu@miami.edu,
ywang393@syr.edu, 
gang.quan@fiu.edu
}

% \author{Zihao Liu}
% \affiliation{%
%   \institution{Florida International University}
% }
% \email{zliu021@fiu.edu}

% \author{Tao Liu}
% \affiliation{%
%   \institution{Florida International University}
% }
% \email{tliu023@fiu.edu}

% \author{Wujie Wen}
% \affiliation{%
%   \institution{Florida International University}
% }
% \email{wwen@fiu.edu}

% \author{Lei Jiang}
% \affiliation{%
%   \institution{Indiana University}
% }
% \email{jiang60@iu.edu}

% \author{Jie Xu}
% \affiliation{%
%   \institution{University of Miami}
% }
% \email{jiexu@miami.edu}

% \author{Yanzhi Wang}
% \affiliation{%
%   \institution{Syracuse University}
% }
% \email{ywang393@syr.edu}

% \author{Gang Quan}
% \affiliation{%
%   \institution{Florida International University}
% }
% \email{gang.quan@fiu.edu}

%%The default list of authors is too long for headers}
%\renewcommand{\shortauthors}{T. Liu et al.}

\begin{abstract}

%\textcolor{red}{abstract here}
As one of most fascinating machine learning techniques, deep neural network (DNN) has demonstrated excellent performance in various intelligent tasks such as image classification. DNN achieves such performance, to a large extent, by performing expensive trainings over huge volumes of training data. To reduce the data storage and transfer overhead in smart resource-limited Internet-of-Thing (IoT) systems, effective data compression is a ``must-have" feature before transferring real-time produced dataset for training or classification. While there have been many well-known image compression approaches (such as JPEG), we for the first time find that a  human-visual based impage compression approach such as JPEG compression is not an optimized solution for DNN systems, especially with high compression ratios. To this end, we develop an image compression framework tailored for DNN applications, named ``DeepN-JPEG", to embrace the nature of deep cascaded information process mechanism of DNN architecture. 
%A new adaptive quantization design that can well trade
%by leveraging the statistical frequency information augmented from a few sampled images.
%JPEG. 
Extensive experiments, based on ``ImageNet" dataset with various state-of-the-art DNNs, show that ``DeepN-JPEG" can achieve $\sim3.5\times$ higher compression rate over the popular JPEG solution while maintaining the same accuracy level for image recognition, demonstrating its great potential of storage and power efficiency in DNN-based smart IoT system design.

\end{abstract}

%\keywords{FPT-Spike, Time coding, SNN, Neuromorphic}
\maketitle

%\vspace{-6pt}
\section{Introduction}
\label{sec:intro}

Pervasive mobile devices, sensors and Internet of Things (IoT) are nowadays producing ever-increasing amounts of data. 
%overwhelming amounts of data. 
%that are capable of driving the development of many new applications. 
The recent resurgence in neural networks---the deep-learning revolution, further opens the door for 
intelligent data interpretation, turning the data and information into actions that create new capabilities, richer experiences and unprecedented economic opportunities. 
%turning data and information into actions that create new capabilities, richer experiences and unprecedented economic opportunities. 
%To better interpret such ``big data'' and unleash its potentials in realistic applications, 
%Thanks to recent machine learning model innovation and computing hardware advancement,
%Indeed, 
For example, deep neural network (DNN) has become the \emph{de facto} technique that is making breakthroughs in a myriad of real-world applications ranging from image processing, speech recognition, object detection, game playing and driver-less cars~\cite{lecun2015deep,szegedy2016overview,silver2016alphago,web6,web5,web7}. 
%Meanwhile, 
%as an example of commercialization attempts of such a technique towards real products, 
%many companies are beginning the commercialization journey, i.e. a class of DNN-based emerging web-service applications for image categorization, face detection and image tagging, such as Google Cloud Vision API, Microsoft Azure Computer Vision API, Amazon Rekognition, Clarifai and IBM Watson~\cite{IBM,micro_API,google_API,clarifai,amazon}. 

The marriage of \emph{big data} and \emph{deep learning} leads to the great success of artificial intelligence, but it also raises new challenges in data communication, storage and computation~\cite{soro2009survey} incurred by the growing amount of distributed data and the increasing DNN model size. For resource-constrained IoT applications, while recent researches have been conducted~\cite{liu2016memristor,han2016eie} to handle the computation and memory-intensive DNN workloads in an energy efficient manner, \textit{there lack efficient solutions to reduce the power-hungry data offloading and storage on terminal devices like edge sensors, especially in face of the stringent constraints on communication bandwidth, energy and hardware resources}. Recent studies show that the latencies to upload a JPEG-compressed input image (i.e. 152KB) for a single inference of a popular CNN--``AlexNet'' via stable wireless connections with 3G (870ms), LTE (180ms) and Wi-Fi (95ms), can exceed that of DNN computation (6$\sim$82ms)  by a mobile or cloud-GPU~\cite{kang2017neurosurgeon}. Moreover, the communication energy is comparable with the associated DNN computation energy. 
%Reducing the data offloading overhead is very essential in edge computing.

%Recent studies show that the delay and energy of data movement from terminal devices to servers can even surpass that of DNN testing computation performed by a GPU~\cite{kang2017neurosurgeon}. 

Data compression is an indispensable technique that can greatly reduce the data volume needed to be stored and transferred, thus to substantially alleviate the data offloading and local storage cost in terminal devices. As DNNs are contingent upon tons of real-time produced data, it is crucial to compress the overwhelming data effectively. Existing image compression frameworks (such as JPEG) can compress data aggressively, but they are often optimized for the Human-Visual System (HVS) or human's perceived image quality, which can lead to unacceptable DNN accuracy degradation at higher compression ratios (CR) and thus significantly harm the quality of intelligent services. As shown later, testing a well-trained \emph{AlexNet} using $CR=\sim5\times$ compressed JPEG images (w.r.t.  $CR=1\times$ high quality images ), can lead to $\sim9\%$ image recognition accuracy reduction for the large scale dataset--- \emph{ImageNet}, almost offsetting the improvement brought by more complex DNN topology, i.e. from \emph{AlexNet} to \emph{GoogLeNet} (8 layers, 724M MACs v.s. 22 layers, 1.43G MACs)~\cite{krizhevsky2012imagenet,szegedy2015going}. This prompts the need of developing an DNN-favorable deep compression framework.

In this work, we for the first time develop a high efficient image compression framework specifically target on DNN, named \textbf\textbf{DeepN-JPEG}. Unlike existing compressions that are developed by taking the human perceived distortions as the top priority, \textbf\textbf{DeepN-JPEG} preserves important features crucial for DNN classification with guaranteed accuracy and compression rate, thus to drastically lower the cost incurred by data transmission and storage in resource-limited edge devices. 
%Our architecture naturally orchestrates the data quality and the inherent data processing capability of the heterogeneous DNN accelerators, thus to significantly lower the associated overhead incurred by data transmission. 
%orchestrates the  
%the unique properties of DNN, i.e. accuracy etc., 
%as the design constraints, 
%we for the first time discovered that the human-visual based JPEG compression is not an optimized solution for DNN systems especially under high compression ratios. 
%We then developed a deep compression framework tailored for DNNs, named ``DeepN-JPEG", to embrace the nature of unique deep cascaded information process mechanism of DNN architecture. 
Our major contributions are: 
%The main contributions of our work can be summarized as:
\begin{enumerate}
\item We propose a semi-analytical model to capture the image processing mechanism differences between human visual system (HVS) and deep neural network at frequency domain;
%Understand and analysis DNN response for training image dataset. 
\item We develop an DNN-favorable feature refinement methodology by leveraging the statistical frequency component analysis of various image classes;
%We propose a "DeepN-JPEG" DNN suitable image compression method which require less memory with no accuracy degrade. 
\item We propose piece-wise linear mapping function to link statistical information of refined features to individual quantization values in the quantization table, thus to optimize the compression rate with minimized accuracy drop.
%The results shown our design can significantly improve storage, power efficiencies and improve performance.
%quantization table design methods
\end{enumerate}
Experimental results show that \textbf\textbf{DeepN-JPEG} can achieve much higher compression efficiency (i.e.$\sim3.5\times$) than that of JPEG solution while maintaining the same accuracy level with the same hardware cost, demonstrating the great potentials for its applications in low-cost and ultra-low power terminal devices, i.e. edge sensors.

\section{Background and Motivation}
\label{pre}
% In this section we present the background of Deep Neural Networks and the data flow at data loading process. Then we take a overview of JPEG image compression algorithm. We finally show the motivation of our work.
\subsection{Basics of Deep Neural Networks }

DNN introduces multiple layers with complex structures to model a high-level abstraction of the data~\cite{hinton2006reducing}, and exhibits high effectiveness in finding hierarchical patterns in high-dimensional data by leveraging the deep cascaded layer structure~\cite{he2016deep,krizhevsky2012imagenet,simonyan2014very,szegedy2015going}. Specifically, 
%The neural network model consists of a comprehensive layer topology and associated parameters (or weights) in each layer. 
%\tao{consider to put more details for neural processing.}
%Fig.~\ref{back_dnn}(a) shows the neural network model and the underlying neural processing in a typical DNN. 
the convolutional layer extracts sufficient feature maps from the inputs by applying kernel-based convolutions, the pooling layer performs a downsampling operation (through max or mean pooling) along the spatial dimensions for a volume reduction, and the fully-connected layer further computes the class score based on the weighted results and non-linear activation functions. Softmax regression (or multinomial logistic regression)~\cite{bishop2006pattern} is usually adopted in the last layer of most DNNs for a final decision.
%handle the final classification of different classes. 

To perform realistic image recognition, the DNN hyper-parameters are trained extensively through an overwhelming amount of input data. 
%an overwhelming amount of input data has involved to . 
For instance, the large-scale dataset--ImageNet~\cite{imagenet_cvpr09}, which consists of 1.3 Million high resolution image samples ($\sim140$ Gigabyte) in 1K categories, is dedicated to training state-of-the-art DNN models for image recognition task.

\subsection{HVS-based JPEG Compression}

\begin{figure}[b]
\begin{centering}
\includegraphics[width=1\columnwidth]{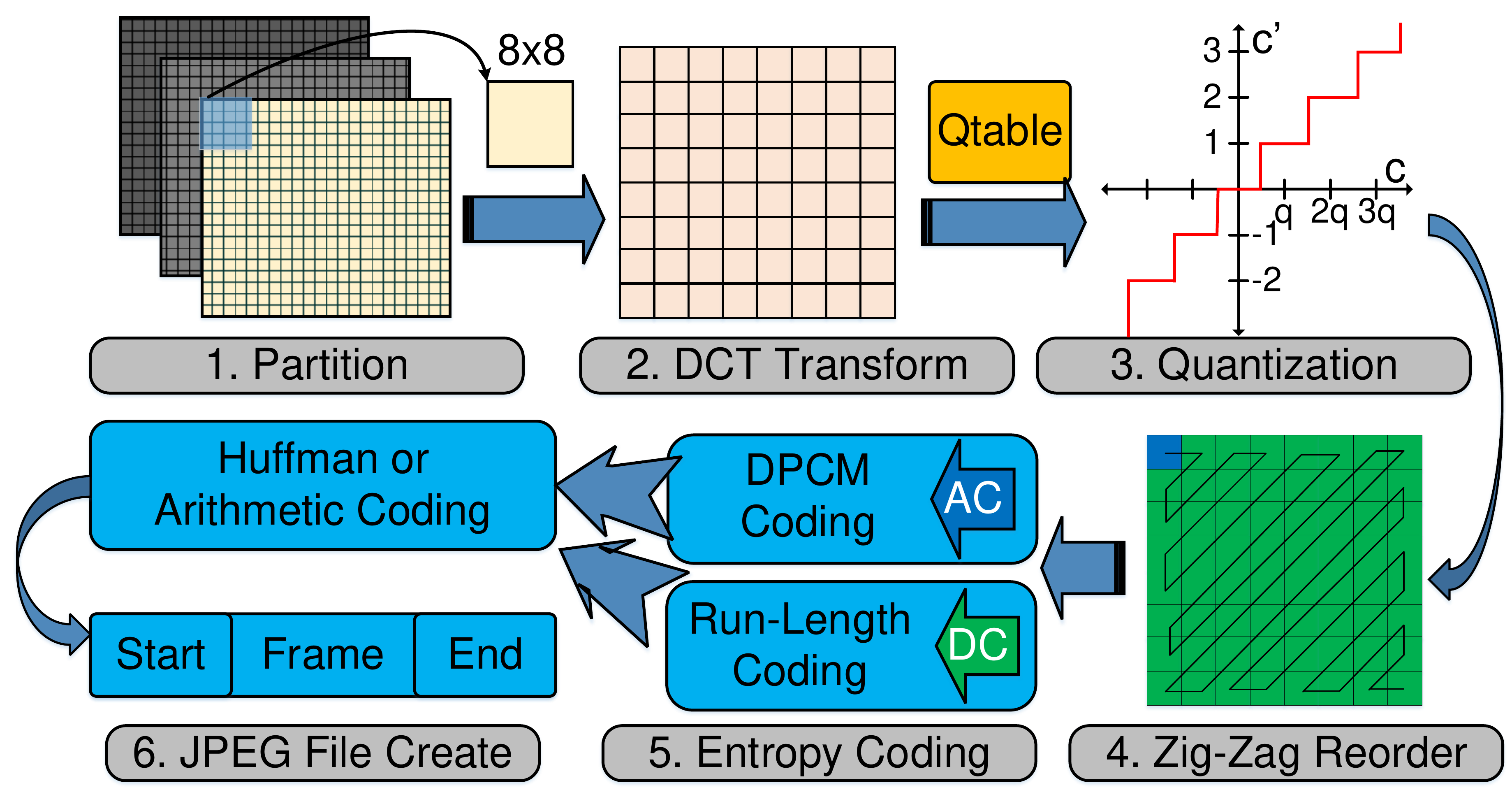}
\end{centering}
\vspace{-10pt}
\caption{Briefly overview of JPEG compression technology .}
\label{JPEGflow}
\vspace{-10pt}
\end{figure}
%In most visual recognition applications, image compression is introduced against the massive data processing. 
It is widely agreed that massive images and videos, as the major context to be understood by deep neural networks, dominate the wireless bandwidth and storage ranging from edge devices to servers. Hence, in this work, we focus on the image compression.

JPEG~\cite{wallace1992jpeg} is one of the most popular lossy compression standards for digital images. It also forms the foundation of most commonly used video compression formats like Motion JPEG (MPEG) and H.264 etc~\cite{ratnakar2000efficient}. 
As shown in Fig.~\ref{JPEGflow}, for each color component, i.e. the RGB channels, the input image is first divided into $8\times8$ non-overlapping pixel blocks, then 2D Fourier Discrete Cosine (DCT) Transform is applied at each $8\times8$ block to generate 64 DCT coefficients $c_{i,j}$, $i\in {0,...,7}$, $j\in {0,...,7}$, of which $c_{0,0}$ is direct current (DC) coefficient, and $c_{0,1},...,c_{7,7}$ are 63 alternating current (AC) coefficients.
Each 64 DCT coefficients are quantized and rounded to the nearest integers as $c'_{i,j}=round[\frac{c_{i,j}}{q_{i,j}}]$,
here $q_{i,j}$ is the individual parameter of the 64-element quantization table provided by JPEG~\cite{wallace1992jpeg}. The table is designed to \textit{preserve the low-frequency components and discard high-frequency details because human visual system (HVS) is less sensitive to the information loss in high frequency bands}~\cite{zhang2017just}. As a many-to-one mapping, such quantization is fundamentally lossy (i.e. $c_{i,j}\neq c'_{i,j} \times q_{i,j}$ at the decompress stage), and can generate more shared quantized coefficients (i.e. zeros) for a better compression. After quantization, all the quantized coefficients are ordered into the ``zig-zag'' sequence following the frequency increasing. Finally, the differential coded DC and run-length coded AC coefficients will be further compressed by lossless Huffman or Arithmetic Coding. Increasing (reducing) the compression ratio (CR) can be usually realized by scaling down (up) the quantization table through adjusting the quantization factor (QF). A larger QF indicates better image quality but a lower CR. A reserved procedure of aforementioned steps can decompress an image. 

\begin{figure}[t]
 \begin{centering}
 \includegraphics[width=1\columnwidth]{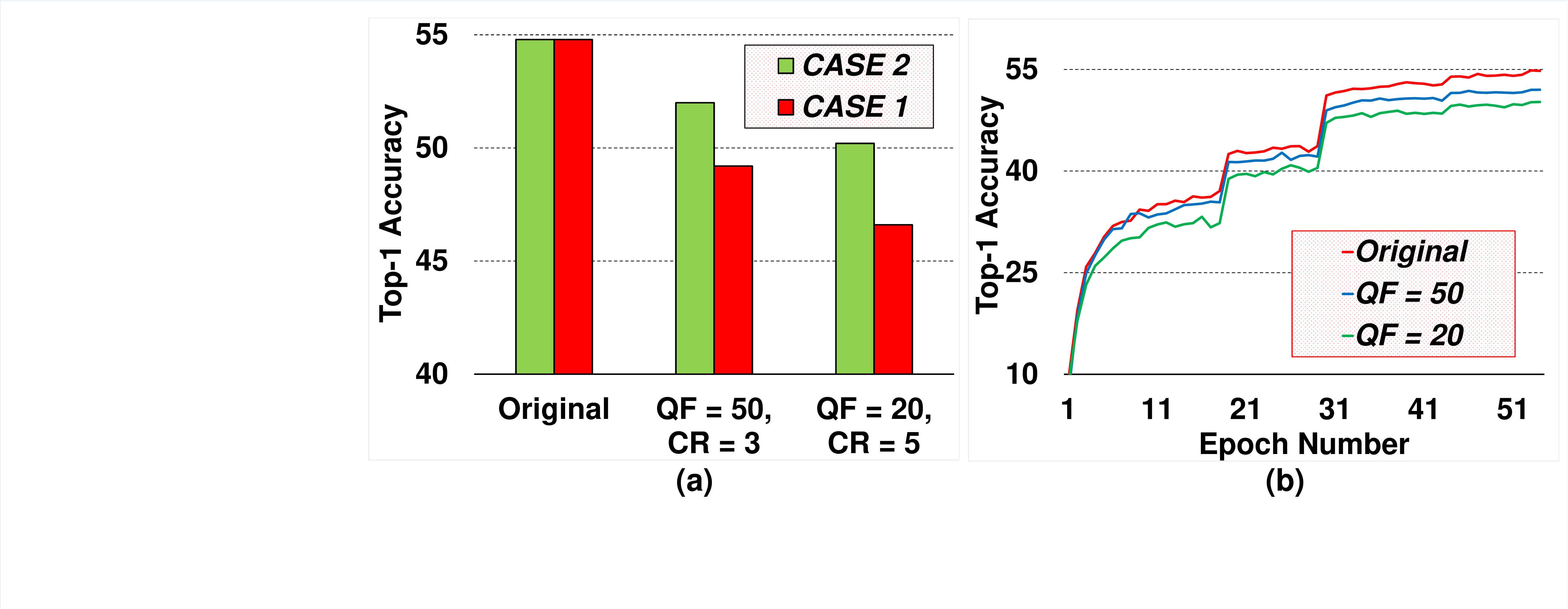}
 \end{centering}
 \vspace{-10pt}
 \caption{ \small (a) Accuracy v.s. JPEG CRs of  ``AlexNet'' for CASE 1/2; (b) CASE 2--Accuracy w.r.t Epoch Number at various CRs.}
 \label{acc}
\vspace{-10pt}
\end{figure}

\subsection{Inefficient HVS Compression for DNNs}
%However in practice, 
\textbf{DNN suffers from dramatic accuracy loss if using existing HVS-based compression techniques to aggressively compress the input images for more efficient data offloading and storage}: To explore how existing compressions can impact the accuracy of DNN, we have conducted following two sets of experiments: \textbf{CASE 1}: training DNN model by high quality JPEG images (QF=100), but testing it with images at various CRs or QFs (i.e. QF=100, 50, 20); \textbf{CASE 2}: training DNN model by various compressed images (QF=100, 50, 20), but testing it only with high quality original images (QF=100). In both cases, a representative DNN example--``AlexNet''~\cite{krizhevsky2012imagenet} with 5 convolutional layers, 3 fully connected layers and 60M weight parameters is trained with the ImageNet dataset for large scale visual recognition. 
%\tao{taking care the QF-CR consistency and consider to measure both cases in evaluation.}
%We use the ``top 1'' accuracy--the classification result (the one with highest probability) must be exactly as the expected answer to measure our results.

%For ``CASE 1", we assume images from the edge sensors are subject to more aggressive compressions due to limited local storage, bandwidth or battery sometime, while the DNN model dedicated to the classification in mobile device or cloud is actually pre-trained with ideal images. ``CASE 2" assumes that ideal high quality testing images can be always guaranteed but the model is trained with lower quality and higher compressed images. 

%We select the state-of-the-art benchmark used in Large Scale Visual Recognition Challenge (ILSVRC)--``ImageNet" for our image classification task~\cite{imagenet_cvpr09}. The dataset includes 1.3M various sized images in 1K categories. A representative DNN example--``AlexNet'' with 5 convolutional layers, 3 fully connected layers and 60M weight parameters is adopted~\cite{krizhevsky2012imagenet}. We use the ``top 1'' accuracy--the classification result (the one with highest probability) must be exactly as the expected answer to measure our results.
%the model answer (the one with highest probability) must be exactly the expected answer
As Fig.~\ref{acc} (a) shows, the ``top-1" testing accuracies characterized from both cases degrade significantly as the CR increases from 1 to 5 (or QF from 100 to 20). To achieve the best CR (QF=20, CR=5), the accuracy of CASE 1 (CASE 2) can be even dropped by $\sim9\%$ ($\sim5\%$) than that of the original one (QF=100, CR=1). Note that the accuracy improvement of ImageNet from ``AlexNet" to ``GoogLeNet" is merely $~\sim9\%$, despite of the significant increased number of layers (8 v.s. 22) and multiply-and-accumulates (724M v.s. 1.43G).
%and $~\sim5\%$, respectively, despite of more complex DNN architectures. 
We also observe that ``CASE 2" can always exhibit smaller accuracy reduction than ``CASE 1" across all CRs ranging from CR=3 to CR=5. This clearly indicates that training the DNN with more compressed JPEG images (compared with testing ones) can slightly alleviate the accuracy dropping, but cannot completely address this issue. As Fig.~\ref{acc} (b) shows, the accuracy gap between a higher CR (or low QF, i.e. QF=20) and the original one (CR=1) for CASE 2 is maximized at the last testing epoch. Apparently, existing compressions like JPEG, which are centered around human visual system, are not optimized solutions for DNNs, especially at a higher compression ratio.

\section{Our Approach}
Developing efficient compression frameworks has been widely studied in applications like image and video processing, however, all these researches are taking the human perceived distortions as the top priority, rather than the unique properties of deep neural networks, such as accuracy, deep cascaded data processing, etc. In this section, we first discover the different views of human visual system and deep neural network in image processing, and then propose the DNN-favorable JPEG-based image compression framework--``DeepN-JPEG".

%In this section, we first explore the difference between human vision system and deep neural network and the underlying design challenges.

%present our design of proposed DNN-favorable JPEG-based image compression framework--the DeepN-JPEG. To design the effective and efficient compression, we first explore the difference between human vision system and deep neural network and the underlying design challenges.

% Our method is a variation version of DNN suitable JPEG without change any JPEG compress steps. The main idea is we smartly modify the quantization table to ensure the image contain all frequency components. And then allow certain intensity of errors on it to guarantee the accuracy. We first overview our design and then discuss how to design the quantization table. 

\subsection{Modeling the difference of HVS and DNN}
\label{Analysis}
% \begin{figure}[t]
% \begin{centering}
% \includegraphics[width=1\columnwidth]{figures/quancompare}
% \end{centering}
% \caption{Original coefficient distribution and after de-quantization at different quantization value coefficient distribution.}
% \label{q_compare}
% \end{figure}

\begin{figure}[t]
\begin{centering}
\includegraphics[width=1\columnwidth]{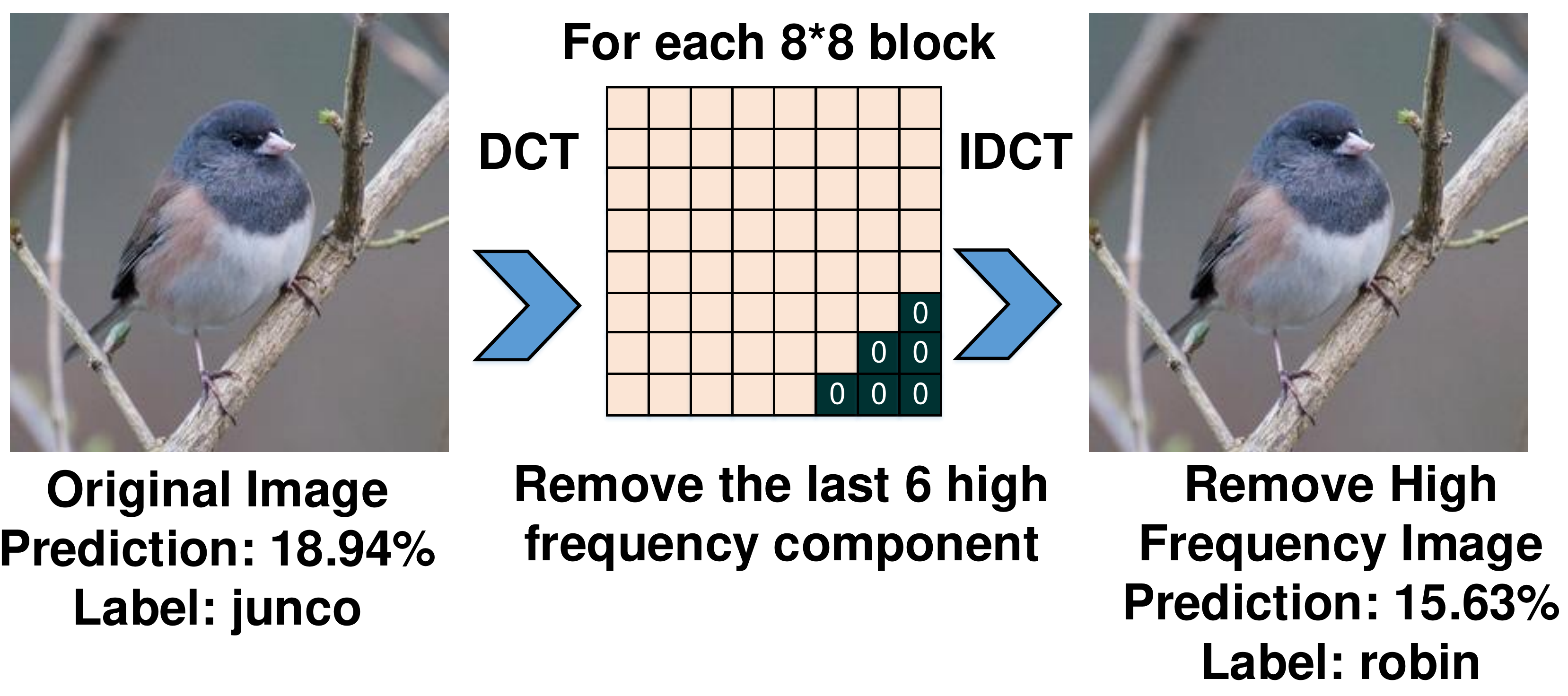}
\end{centering}
\vspace{-10pt}
\caption{Feature degradation will impact the classification.}
\label{hfrm}
\vspace{-10pt}
\end{figure}

We have initialized our studies on an interesting problem: \textit{What are the major differences of image processing between human vision system (HVS) and deep neural network?} This should help on explaining the aforementioned accuracy reduction issue, thus to guide the development of DNN-favorable compression framework. \textit{Our observation is that DNNs can response to any important frequency component precisely, but human visual system focuses more on the low frequency information than high frequency ones, indicating fewer features to be learned by DNNs after the HVS-inspired compression.} 
Assume $x_k$ is a single pixel of a raw image \textbf{X}, and $x_k$ can be represented by $8\times8$ DCT in JPEG compression: 
\begin{equation}
x_k=\sum_{i=0} ^{i=7}\sum_ {j=0} ^{n=7} c_{(k,i,j)}\cdot b_{(i,j)}
\end{equation}
where $c_{(k,i,j)}$ and $b_{(i,j)}$ are the DCT coefficient and corresponding basis function at 64 different frequencies, respectively.
Because the human visual system is less sensitive to high frequency components, a higher CR can be achieved in JPEG compression by intentionally discarding the high frequency parts, i.e. zeroing out the associated DCT coefficient $c_{(k,i,j)}$ through scaled quantization. On the contrary, DNNs examine the importance of the frequency information in a quite different way. The gradient of the DNN function $F$ with respect to a basis function $b_{(i,j)}$ can be calculated as:
\begin{equation}
{ \frac{\partial F}{\partial b_{(i,j)}}=\frac{\partial F}{\partial x_k}\times\frac{\partial x_k}{\partial b_{i,j}}=\frac{\partial F}{\partial x_k}\times c_{(k,i,j)} }
\label{basis}
\end{equation}
Eq.~\ref{basis} implies that the contribution of a frequency component ($b_{i,j}$) of a single pixel $x_k$ to the DNN learning will be mainly determined by its associated DCT coefficient ($c_{(k,i,j)}$) and the importance of the pixel ($\frac{\partial F}{\partial x_k}$). Here $\frac{\partial F}{\partial x_k}$ is obtained after the DNN training, while $c_{(k,i,j)}$ will be distorted by the image compression (i.e. quantization) before training. 
If $c_{(k,i,j)}=0$, the frequency feature ($b_{i,j}$), which may carry important details for DNN feature map extraction, cannot be learned by DNN for weights updating, causing a lower accuracy.

It is often the case in a highly compressed JPEG image, given that $c_{(k,i,j)}$s of high frequency parts (usually small in nature images) are quantized to zero to ensure a better compression rate. 
% Fig.~\ref{q_compare} shows the quantization induced feature degradation. With a moderate CRatio, i.e. QF = 80, the de-quantized coefficient distribution is similar to the original one but little sparse. However increasing the CRatio, i.e. QF = 60/20, is resulting in the over-sparse coefficients. 
As a result, DNNs can easily misclassify aggressively compressed images if their original versions contain important high frequency features. In CASE 1 (see Fig.~\ref{acc}(a)), the DNN model trained with original images learns comprehensive features, especially high frequency ones that are important in some images, however, such features are actually lost in some more compressed testing images, causing considerable misclassification rate. Fig.~\ref{hfrm} demonstrates such an example--the ``junco" is mis-predicted as ``robin" after removing the top six high frequency components, despite that the differences are almost indistinguishable by human eyes. In CASE 2 (see Fig.~\ref{acc}(b)), the model is trained to make decisions solely based on the limited number of features learned from more compressed training images, and the additional features in high quality testing images cannot be detected by DNN for accuracy improvement.

% In this section we describe the proposed DeepN-JPEG method. The main idea of our method is to redesign a DNN suitable quantization table based on frequency components statistical information of training dataset. Our design has three main components: 1. Image complexity-aware sampling, 2. frequency component analysis and 3. quantization table generation. 

\begin{figure*}[t]
\begin{centering}
\includegraphics[width=\textwidth]{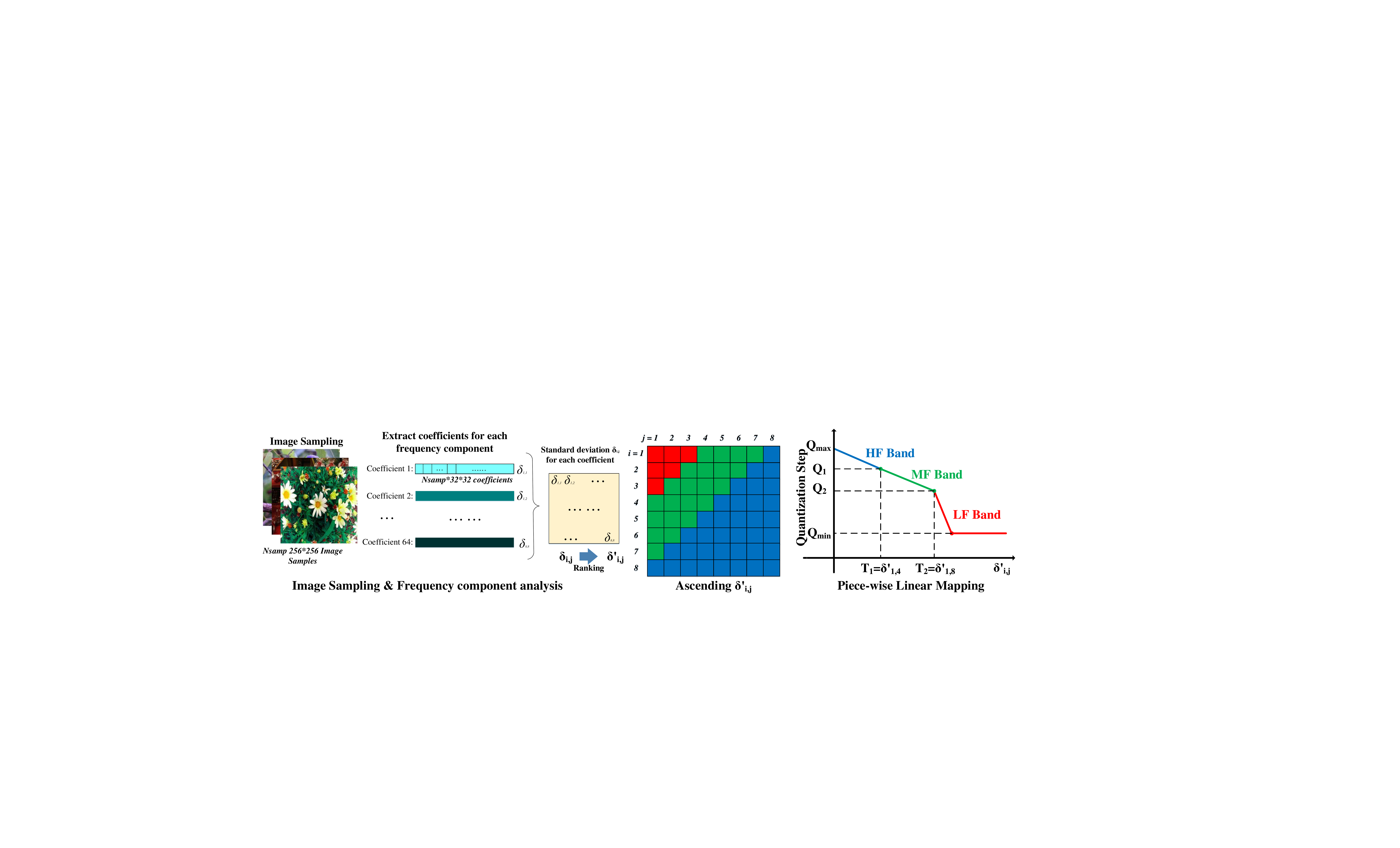}
\end{centering}
\vspace{-10pt}
\caption{An overview of heuristic design flow of ``DeepN-JPEG" framework.}
\label{flowchart}
\vspace{-10pt}
\end{figure*}

\subsection{DNN-Oriented DeepN-JPEG Framework}
%Based on aforementioned observation, 
To develop the ``DeepN-JEPG" framework, it is essential to minimize the distortion of frequency features that are most important to DNN, thus to maintain the accuracy as much as possible.
%Similar to developing human visual system based compression
%techniques, 
%our first task is to minimize the distortion of frequency features that are most important to the DNN. 
%As Sec. 3.1.3 shows, 
As quantization is the principle factor to cause important feature loss, i.e. removing less significant high frequency parts by using a larger quantization step in JPEG, the key step of ``DeepN-JEPG" is to re-design such HVS-inspired quantization table to be DNN favorable, i.e. achieving a better compression rate than JPEG without losing needed features.
%To well balance the CRatio and the needed features, a
%feasible solution is to re-design the HVS-inspired quantization table to be DNN favorable.
%Based on aforementioned observations, 
%we intend to develop an DNN favor-able compression to size down the data storage and data traffic by re-designing the HVS-inspired quantization table to be DNN favorable.  
Although the quantization table redesign has been proved to be a feasible solution in various applications, such as feature detection~\cite{chao2013design}, visual search~\cite{duan2012optimizing}, it is an intractable optimization problem for ``DeepN-JPEG" because of the complexity of parameter searching~\cite{hopkins2017simulated}, and the difficulty of a quantitative measurement suitable to DNNs.
%difficulty of measurement the distortion index suitable for DNNs. 
For example, it is non-trivial to characterize the implicit relationship between image feature (or quantization) errors and DNN accuracy loss. Moreover, the characterized results could vary according to the DNN structure. Therefore, it is very challenging to develop a generalized DNN-favorable compression framework. 

Our analysis in section~\ref{Analysis} indicates that the contribution of a frequency band to DNN learning is strongly related with the magnitude of the band coefficient. Inspired by this key observation, our ``DeepN-JEPG" is developed upon a heuristic design flow (see Fig.~\ref{flowchart}): 
1) Sample representative raw images from each class and further characterize the importance of each frequency component through frequency analysis on sampled sub dataset; 2) Link the statistical information of each feature with the quantization step of quantization table through proposed ``Piece-wise Linear Mapping''. 

\subsubsection{\textbf{Image Sampling and Frequency Component Analysis}}
In ``DeepN-JPEG" framework, our first step is
%the first task in our design is to %conduct the complexity-aware sampling on image dataset for quantifying the input features. 
to sample all classes within the labeled dataset, for more comprehensive feature analysis. To extract the representative features from the whole dataset and rank the importance of those features to DNN, we 
%may gathering more accurate information on features, however it will also introduce the expensive time and computation requirements. On the other hand, sampling an inadequate sub-dataset may not capture the enough overall features for designing the representative quantization table thus insufficient compressing.
% \begin{algorithm}[b]
% \small
% \caption{\label{alg_1} Image Sampling Algorithm}
% \DontPrintSemicolon
% \tcp{Pseudocode of Image Sampling Algorithm}%: $V_{max}$/$T_{max}$ is recorded.}
% Input: Full dataset;\;
% C: \# of Classes;\;
% N: \# of images in each class;\;
% k: Interval for sampling images;\;
% Spath: Path of Sampled Images;\;
% Nsamp: \#number of sampled images;\;
% Output: Spath, Nsamp.\;
% \ForEach{class $class_i$ in [$class_1$ .. $class_C$]}{
%    m = 0; \tcp{count the number of images in certain class}
%    \ForEach{image $img_j$ in [$img_1$ .. $img_N$]}{m++;\;
%      \If{m \% k = 0 }{  Spath record (Path of $img_j$)
%      }
%  }
% }
% Nsamp = \# of pathes in Spath\;
% return Spath \tcp{All the Path of Sampled Images} 
% return Nsamp
% \end{algorithm}
% %\tao{zihao, please update the alg.1 for sampling on all classes.}
%Following aforementioned design constraint, 
%our approach will conduct an overall sampling to cover the more effective features shared among the whole dataset. 
%Consider that the images are usually stored in compressed format rather than raw data. The size of an image file naturally 

implied the feature complexity of the image--smooth image with simple features will be compressed at small size while large size indicates the image consists of more complex features. 
%In order to get comprehensive information of the images at different situation and background. We will select the samples with 
%different size with size 
%index interval $S$ for each class. Later, these representative samples will be preserved for frequency component analysis.
%One approach is to 
characterize the un-quantized DCT coefficient distribution at each frequency band, since the distribution represents the energy of a frequency component~\cite{Rei:TC1983}. Previous studies~\cite{Rei:TC1983} have proven that the un-quantized coefficient can be approximated as normal (or Laplace) distribution with zero mean but different standard deviations ($\delta_{i,j}$). A larger $\delta_{i,j}$ indicates more energy in the band $(i,j)$, hence more contributions to the DNN feature learning. 
%The frequency component analysis is proposed to gather such a statistical frequency distribution for designing the appropriative quantization table for effective compressing.
As shown in algorithm~\ref{alg_2}, each sampled image will be first partitioned into $Nblock$ $8\times8$ blocks, followed by a block-wise DCT. After that, the DCT coefficient distribution at each frequency band will be characterized by sorting all coefficients within the same frequency band across all image blocks collected from different classes of the image dataset. The statistical information, such as the standard deviation $\delta_{i,j}$ of each coefficient, will be calculated based on each individual histogram. Note that such a frequency refinement procedure can precisely tell out the most significant features to DNN, and is different from the simple assumption that low frequency part is always more important than the high ones can easily lead to the DNN accuracy reduction.

\begin{algorithm}[b]
\footnotesize
\caption{\label{alg_2}Frequency component analysis Algorithm}
\DontPrintSemicolon
%\tcp{Pseudocode of Frequency component analysis Algorithm}%: $V_{max}$/$T_{max}$ is recorded.}
C: \# of Classes;\;
N: \# of images in each class;\;
k: Interval for sampling images;\;
Spath: Path of Sampled Images;\;
Nsamp: \#number of sampled images;\;
$fimg_i$: Image in frequency domain;\;
$fc_k$: Frequency components;\;
Nblock: \# of 8*8 blocks after block-wise DCT;\;
$\delta_k$: standard deviation of kth frequency components;\;
\ForEach{class $class_i$ in [$class_1$ .. $class_C$]}{
   m = 0; \tcp{count the number of images in certain class}
   \ForEach{image $img_j$ in [$img_1$ .. $img_N$]}{m++;\;
     \If{m \% k = 0 }{  Spath record (Path of $img_j$)
     }
 }
}
\ForEach{image $Spath$ in [$img_1$ .. $img_{Nsamp}$]}{
   $fimg_i$ = 8*8 block-wise DCT ($img_i$)\;
   \ForEach{$Block_{i,j}$ in [1 .. Nblock]} {
    $Block_{i,j}$ = $fimg_i$[j*8-8:j*8][j*8-8:j*8]\tcp{ith sampled image jth 8*8 block}
      \ForEach{$fc_k$ in [1 .. 64]}{
         $fc_k$ store $Block_{i,j}[k]$\tcp{ith sampled image jth 8*8 block kth frequency component}
    }
  }  
}
\tcp{Statistical Analysis}
\ForEach{$fc_k$ in [1 .. 64]}{
calculate standard deviation $\delta_k$}
return $\delta_k$ \tcp{standard deviation of each frequency components}
\end{algorithm}

% \begin{figure}[b]
% \begin{centering}
% \includegraphics[width=0.8\columnwidth]{figures/map}
% \end{centering}
% \caption{Piece-wise Linear Mapping of DCT standard deviation v.s. quantization step.}
% \label{qscale}
% \end{figure}

\subsubsection{\textbf{Quantization Table Design}}
\label{qtable}
Once the importance of frequency band to DNN is identified by our calibrated DCT coefficient standard deviation, our next question becomes how to link those information to the quantization table design to achieve a higher compression rate with minimized accuracy reduction. The basic idea is to introduce less (more) quantization errors at the critical (less critical) band by leveraging the intrinsic error resilience property of the DNN. To introduce nonuniform quantization errors at different frequency bands, we develop a piece-wise linear mapping function (PLM) to derive the quantization step of each frequency band from the associated standard deviation:  
%Then a nonuniform Piece-wise Linear mapping (PLM) to derive the quantization steps of each frequency band can be expressed as: 
\begin{equation}
\label{plm}
Q_{i,j}=\begin{cases}a-k_1*\delta_{i,j}  & \delta_{i,j} \leq T_1\\b-k_2*\delta_{i,j} &T_1<\delta_{i,j}\leq T_2\\c-k_3*\delta_{i,j} & \delta_{i,j}>T_2 \end{cases}~,~s.t.~ Q_{i,j}\geq Q_{min}
\end{equation}
where $Q_{i,j}$ is the quantization step at the frequency band $(i,j)$. $Q_{min}$ is the lowest quantization step. $a$, $b$, $c$, $k_1$, $k_2$, $k_3$ are fitting parameters. $T_1$ and $T_2$ are thresholds to categorize the 64 frequency bands according to the $\delta^{'}_{i,j}$, i.e. ascending order of the magnitude of $\delta_{i,j}$. As right part of Fig.~\ref{flowchart} shows, following the similar frequency segmentation in~\cite{kaur2011steganographic}, the 64 frequency components are divided into three bands: \textbf{Low Frequency (LF)}--1-6 frequency components (largest $\delta^{'}_{i,j}$), \textbf{Middle Frequency (MF)}--7-28 and \textbf{High Frequency (HF)}--29-64 (smallest $\delta^{'}_{i,j}$). Hence, we adopt $T_1 = \delta^{'}_{1,8}$ and $T_2 = \delta^{'}_{1,4}$ in our design. Three different slopes--$k_1$, $k_2$, $k_3$, are assigned to HF band, MF band and LF band, respectively. 

\section{Design Optimization}

\begin{figure}[t]
\begin{centering}
\includegraphics[width=1\columnwidth]{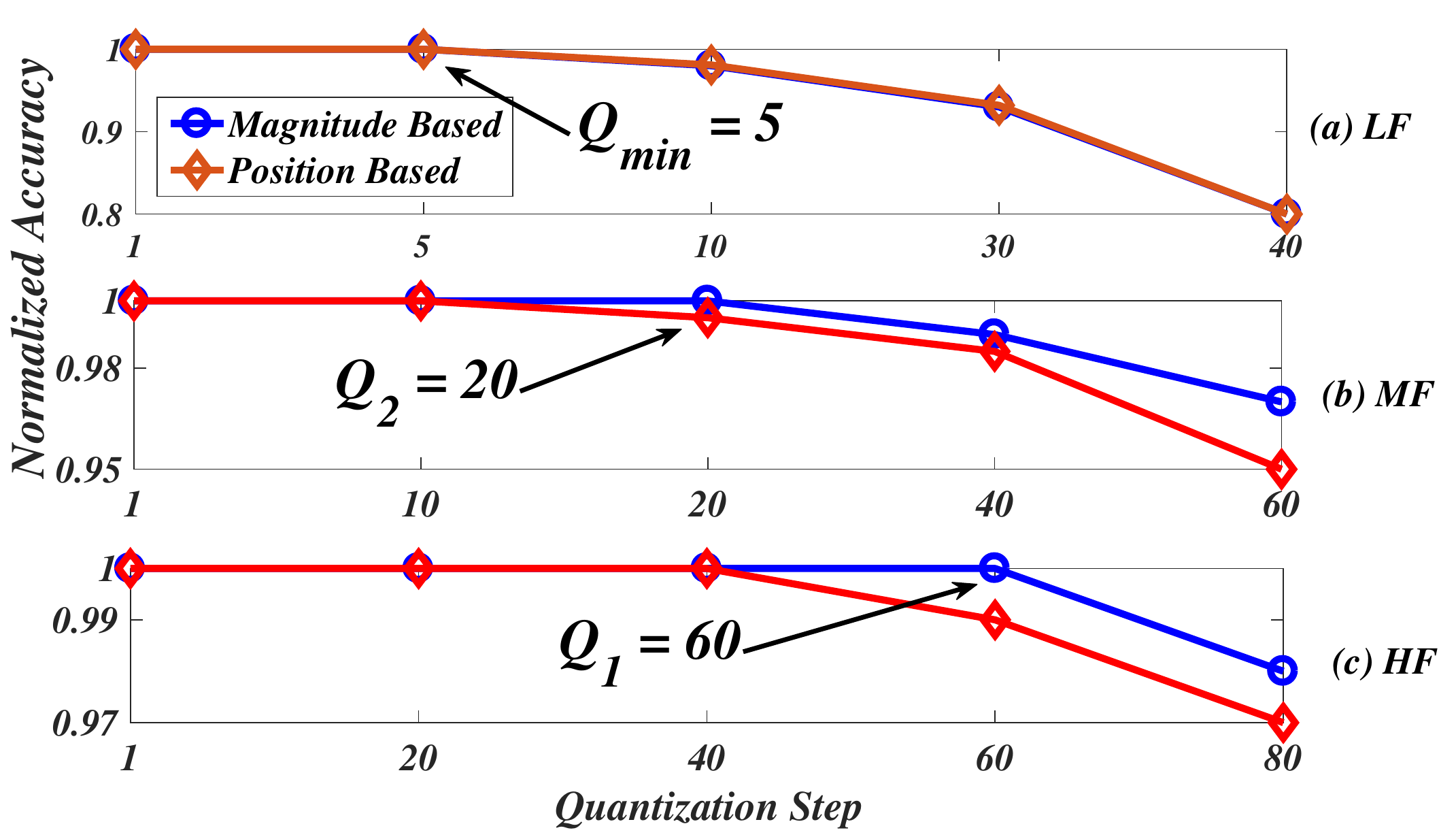}
\end{centering}
\vspace{-10pt}
\caption{Parameter optimization for different frequency bands.}
\label{band}
\vspace{-10pt}
\end{figure}

\begin{figure}[b]
\begin{centering}
\includegraphics[width=1\columnwidth]{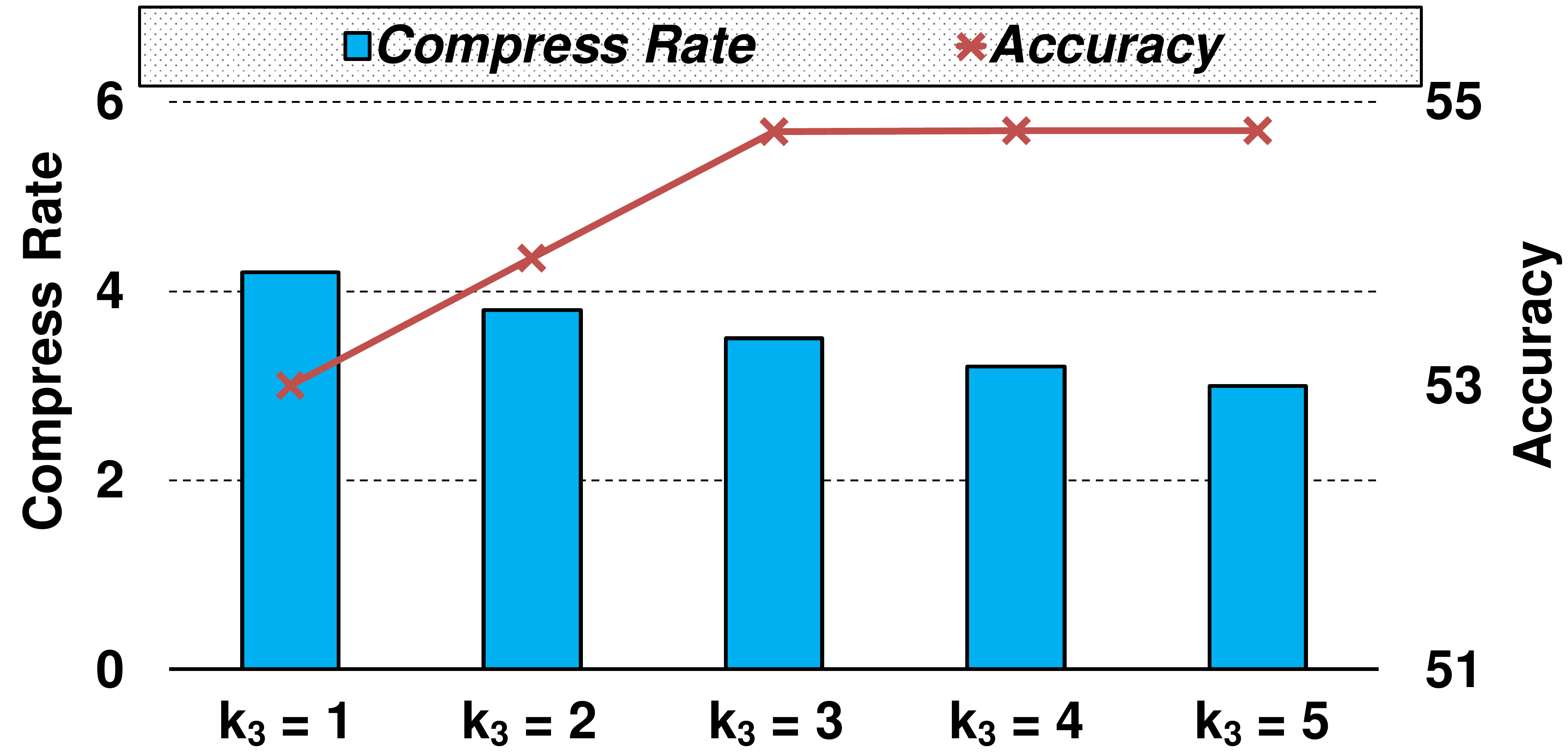}
\end{centering}
\vspace{-10pt}
\caption{Optimization of $k_3$ parameter in PLM.}
\label{figplm}
% \vspace{-10pt}
\end{figure}

%We will discuss $k_1$, $k_2$ and $k_3$ value in next section. As long as we have $k_1$, $k_2$, $k_3$, $T_1$, $T_2$ and set $a=Q_{max}$ the parameter $b$ and $c$ can be calculated accordingly.

In this section, we explore the parameter optimization for our proposed Piece-wise Linear Mapping based quantization table design. In order to set optimized parameters of Eq.~\ref{plm}, i.e. $k_1$, $k_2$ and $k_3$, we first study the sensitivity of quantization steps to DNN accuracy across the LF, MF and HF bands. 
%the correlations between quantization step and accuracy have been explored for each frequency band. 
% \textbf{Error Tolerance.} 
%For our PLM design the frequency components are divided into three frequency band, in order to make a more accurate relationship between STD and quantization step we need carefully setting the parameters in ~\ref{plm}. In the same frequency band we assume all the frequency components have same linear relationship between STD and quantization steps. So we start with evaluating the most error tolerance capability for each frequency band. We call our frequency band assignment as magnitude based (MB) which is determined by STD and JPEG used assignment as position based (PB). We compare the error tolerance capability of those two methods to evaluate the efficient of our design.
We define our proposed band allocation in ``DeepN-JPEG" as the ``magnitude based'', i.e. to segment the frequency band into three types (LF/MF/HF) according to the magnitude of standard deviation of DCT coefficient. For comparison purpose, we also implement the coarse-grained band assignment method based on its position within a default JPEG quantization table, namely ``position based". We conduct the simulations by only varying the quantization steps of interested frequency bands, while all the others are assigned with minimized quantization steps, i.e. $Q_{i,j}=1$ without introducing any quantization errors.  
%the ``position based'' (PB) band assignment is adopted in the original JPEG. 
%During the analysis on each frequency band, all other frequency bands have been maintained as uncompressed. 
%And using the decompress dataset to training the DNN model to evaluate the error tolerance for this band. We select 4 carefully chosen quantization steps for each frequency band that is: 5, 10, 30 and 40 for LF, 10, 20, 40 and 60 for MF and 20, 40, 60, 80 for HF. 

\textbf{Frequency Band Segmentation.} As Fig.~\ref{band} shows, ``magnitude based" method can always achieve better accuracy than that of ``position based" in both MF and HF bands as the quantization step increases. Moreover, our solution can provide a larger quantization step in both MF and HF bands without accuracy reduction, i.e. 40 v.s. 60 in HF band, which can translate into a higher compression rate than that of JPEG. Besides, we also observe that DNN accuracy starts to drop if $Q_{i,j}>5$ at the LF band, which indicates that statistically the largest DCT coefficients are most sensitive to quantization errors, thus we set $Q_{min} = 5$ as the lower bound of quantization value to secure the accuracy (see Fig.~\ref{band} (a)). Similarly, based on the critical points of Fig~\ref{band} (b) and (c), we can further obtain the quantization steps at the point $T1$ and $T2$, thus to determine the parameters such as $k_1$, $k_2$, $a$ and $b$.

%Besides,        

%The DNN is very sensitive to LF (a), when the quantization step is larger than 5 the accuracy will degrade. 
%From our previous analysis the DNN is more sensitive to the larger coefficients, thus the accuracy is mainly depending on the largest coefficient and the associated quantization error tolerance.
%thus for each frequency band the largest quantization step this band can achieve without accuracy drop is dependents on the largest coefficient in this band. 
%If the largest coefficients can tolerance quantization step equal 5 error then the other coefficients which are smaller will also induce no accuracy drop. 
%Therefore, as shown in Fig.~\ref{band}, we can set the $Q_{min} = 5$ as the lower bound of quantization value for securing the accuracy. 
%Another observation is the MB and PB almost have the same accuracy degradation trend, which imply 
%In our magnitude based assignment, the largest coefficient will be always at the first 6 frequency component. 
%Similarly, to combine the critical point without accuracy degradation in Fig~\ref{band} (b) and (c) with Eq.~\ref{plm}, we can further get the remaining parameters. For example the point 20/60 in (b)/(c) is the smallest/largest quantization step in MF which corresponding to $T_2$/$T_1$ in Fig~\ref{qscale}.Those two points both satisfy the MF linear function in Eq.~\ref{plm} and we can easily calculate $b$ and $k_2$. 
%The parameters in HF linear function also can be obtained by combine setting $a=Q_{max}$ where $Q_{max}=255$ which is the largest quantization step defined in JPEG and $T_1$ point. 

\textbf{Tuning $k_3$ in LF Band.} Unlike the parameters in MF and HF bands, the optimization of $k_3$ in LF band is non-trivial because of its significant impact to accuracy and compression rate. Since $k_3$ cannot be directly decided according to the lower bound $Q_{min}$ and $c$, we 
%However for the LF band, 
%the adjustable parameter $k_3$ can not directly decided according to the lower bound $Q_{min}$ and $c$. To estimate an appropriate value, 
investigate the correlation between compress rate and accuracy based on a variety of $k_3$. As shown in Fig.~\ref{figplm}, a smaller $k_3$ can offer a better compression rate by slightly sacrificing the DNN accuracy. Based on our observation, we choose $k_3=3$ to maximize the compression rate while maintaining the original accuracy.

\section{Evaluation}
% \tao{1. qtable design modify parameters , get the best result compare plm and im
% 2. accuracy and crate combined figure
% 3. fig 10 different structure
% 4. power}

%\subsection{Experimental Setup}
Our experiments are conducted on the deep learning open source framework Torch~\cite{torch}. The ``DeepN-JPEG" framework is implemented by heavily modifying the open source JPEG framework~\cite{IJG}.  
%In our evaluation, extensive experiments are conducted in the deep learning platforms like Torch~\cite{torch} 
%and the Independent JPEG Group~\cite{IJG} to measure the improvements of compression rate, and power consumption on proposed DeepN-JPEG framework. 
The large-scale ImageNet~\cite{imagenet_cvpr09} dataset is adopted to measure the improvement of compression rate and classification accuracy. Specifically, all images are maintained as their original scales in our evaluation without any speed-up trick such as resize or pre-processing. The optimized parameters of ``DeepN-JPEG" framework dedicated to ImageNet are as follows: $a =  255,~b = 80,~c = 240,~T_1 = 20,~T_2 = 60,~k_1= 9.75,~k_2 = 1,~k_3 = 3$.
%To conduct a comprehensive evaluation for practical intelligent applications,
Four state-of-the-art DNN models are evaluated in our experiment--AlexNet~\cite{krizhevsky2012imagenet}, VGG~\cite{simonyan2014very}, GoogLeNet~\cite{szegedy2015going} and ResNet~\cite{he2016deep}. 
%Table~\ref{} shows the specs of all selected data benchmarks and DNN baselines.
%In our experiment, we modify the IJG open-source software to achieve different compress methods. The CIFAR-10 benchmark is trained and tested with VGG baseline while ImageNet benchmark is adopted in evaluations on both ALexNet and GoogleNet baselines. 
% Three DeepN-JPEG candidates are designed in our evaluations, i.e. ``DeepN-L'' (Linear mapping), ``DeepN-S'' (Segmented mapping) and ``DeepN-PL'' (Piece-wise Linear mapping), to compare with several default JPEG candidates with different quality controls, i.e. "c20/c40/c60/c80" with standard JPEG compression quality $20\%\sim80\%$.
% \textcolor{red}{Dr Wen: What is the exact relationship between QF value and compression rate? You should explain it at the very beginning}

\begin{figure}[t]
\begin{centering}
\includegraphics[width=1\columnwidth]{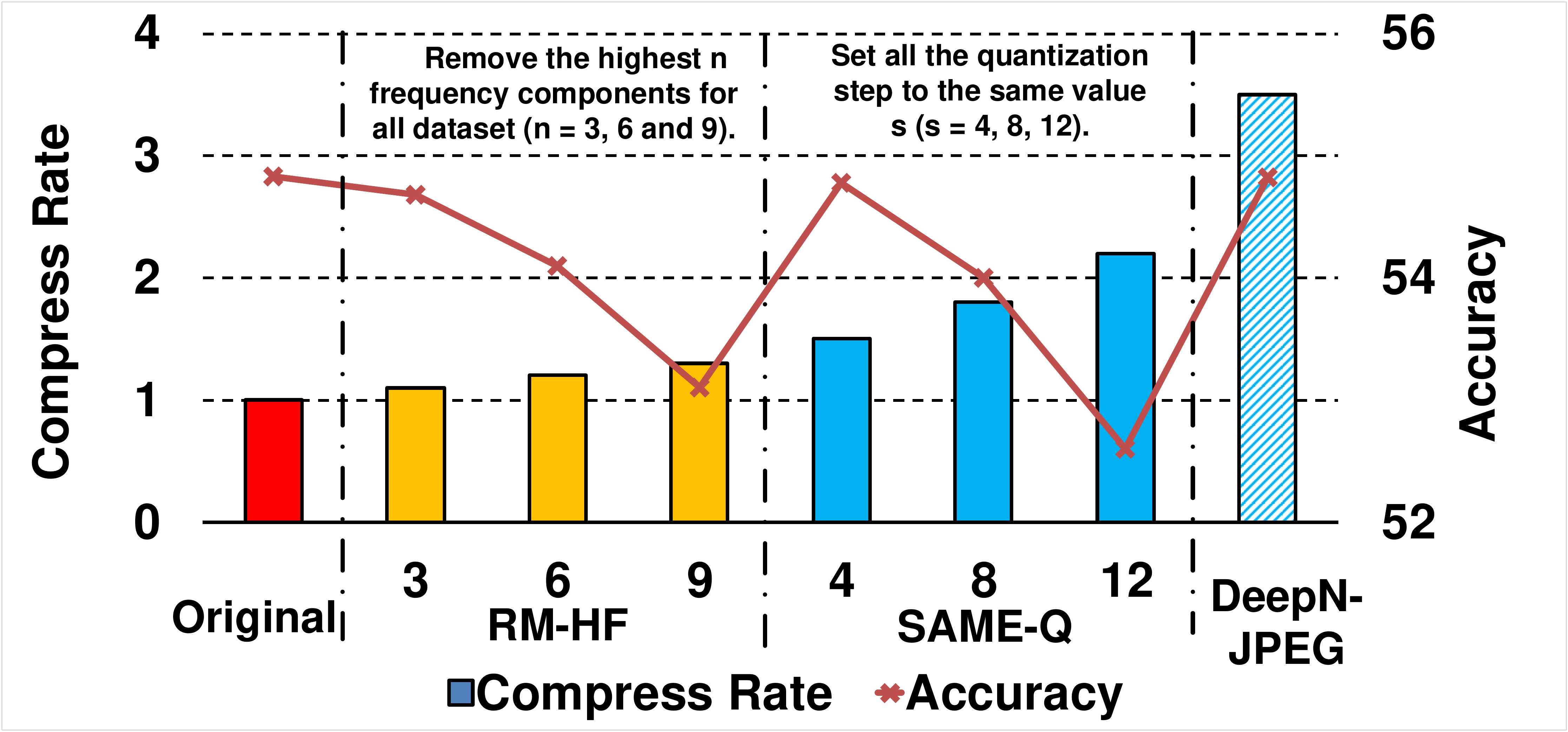}
\end{centering}
\vspace{-10pt}
\caption{The compress rate and  accuracy for different methods.}
\label{compIN}
\vspace{-10pt}
\end{figure}

\begin{figure}[b]
\begin{centering}
\includegraphics[width=1\columnwidth]{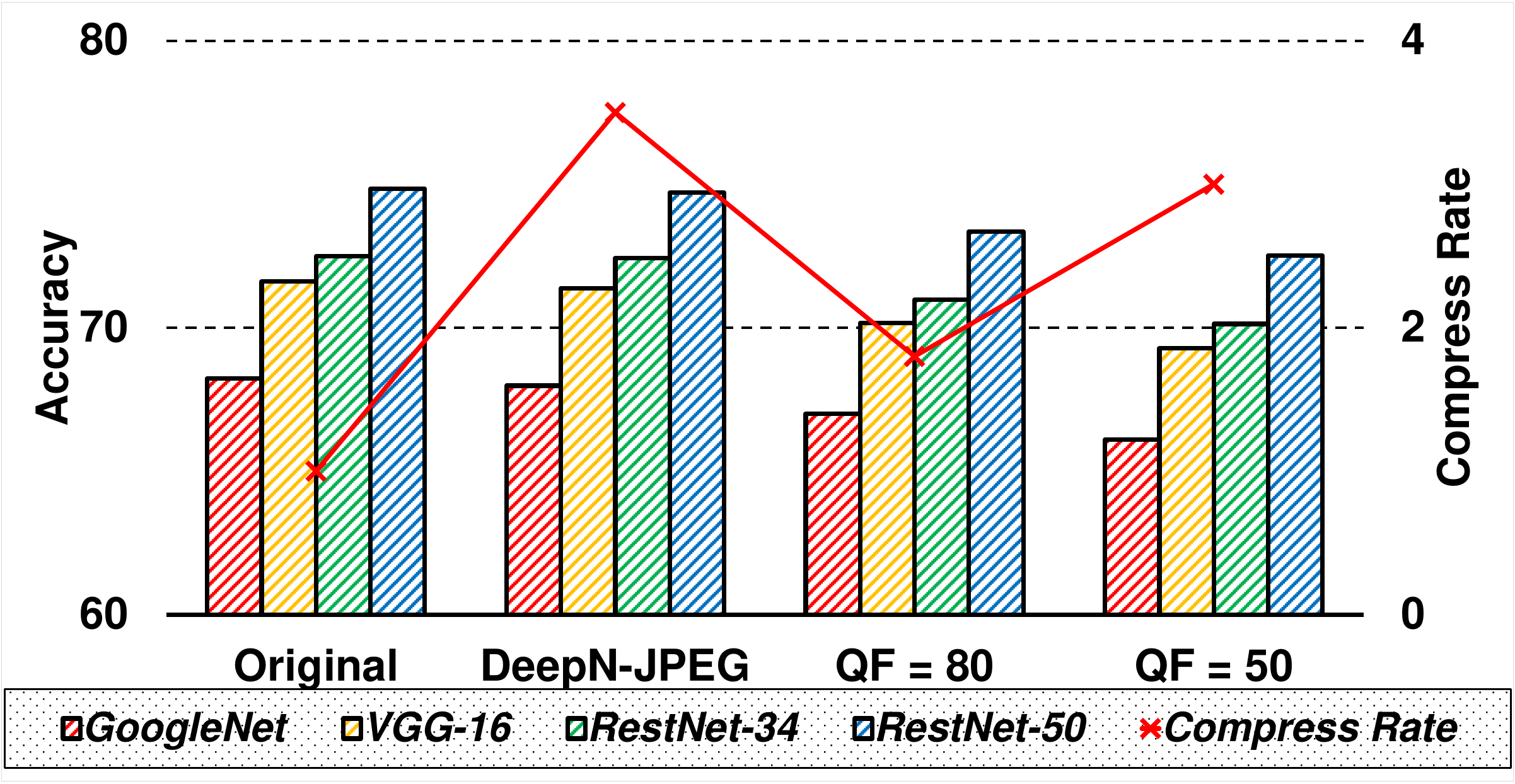}
\end{centering}
\vspace{-10pt}
\caption{The compress rate and  accuracy for different DNN models.}
\label{ac}
% \vspace{-10pt}
\end{figure}

\subsection{Compression Rate and Accuracy}
We first evaluate the compression rate and classification accuracy of our proposed DeepN-JPEG framework. 
%by measuring the compression rate and classification accuracy. 
Three baseline designs are implemented for comparison purpose: the ``original" dataset compressed by JPEG (QF=100, CR=1), ``RM-HF'' compressed dataset and ``SAME-Q'' compressed dataset. Specifically, ``RM-HF'' is extended from JPEG by removing the \textit{top-N high frequency components} from the quantization table to further improve the compression rate, and ``SAME-Q'' denotes a more aggressive compression method with the same quantization step $Q$ for all frequency components.

Fig.~\ref{compIN} compares the compression rate and accuracy based on the ``ImageNet'' dataset  ``AlexNet'' DNN model for all selected candidates. Compared with the ``original", ``RM-HF'' slightly increases the compression rate ($\sim1.1\times-\sim1.3\times$) by removing more highest frequency components (top-3--top-9),  while ``SAME-Q'' achieves better compression rates ($\sim1.5\times-\sim2\times$). However, both schemes suffer from increased accuracy reduction (w.r.t. ``original'') as long as the compression rate becomes higher. On the contrary, our ``DeepN-JPEG" delivers the best compression rate (i.e. $\sim3.5\times$) while maintaining the similar high accuracy as that of original dataset, indicating a promising solution to reduce the cost of data traffic and storage of edge devices for deep learning tasks. 
%for both ``RM-HF'' and ``SAME-Q'', such an ``intentionally data lossy'' leads to considerable accuracy degradations, %thus harming the performance of those ``accuracy-critical'' recognition tasks. 

%In contrast, our ``DeepN'' achieves the best compression rate (i.e. $\sim3.5\times$) among all candidates while still maintaining the original accuracy as that of original dataset, 
%thus a promising solution to improve the performance of data storage and data traffic for DNNs.

\textbf{Generality of DeepN-JPEG.} We also extend our evaluations across several state-of-the-art DNNs to study how the ``DeepN-JPEG" framework responses to different DNN architectures, including GoogLeNet, VGG-16, ResNet-34 and ResNet-50. 
%validate the generality of our proposed ``DeepN-JPEG", We also validate 
%our evaluations are further extended to cover more mainstream DNN models. 
As shown in Fig.~\ref{ac}, our proposed ``DeepN-JPEG" can always maintain the original accuracies (w.r.t. ``Original'') for all selected DNN models.
%achieve the best accuracy among all selected DNN models. 
Although JPEG can easily achieve a similar compression rate as that of ``DeepN-JPEG" by largely reducing the JPEG QF value, e.g. $QF\leqslant50$,  
%can reach to the same compression level as DeepN,  
such an aggressive ``data lossy" compression results in significant side effect on the classification performance of all selected DNN models. In contrast, ``DeepN-JPEG" can preserve both high compression rate and accuracy for all DNNs, thus a generalized solution.

%Therefore, ``DeepN-JPEG" demonstrates the generality 
%such as ``GoogLeNet'', ``VGG'' and ``ResNet''. However, our solution will not impact the original accuracies on these DNN models. 
%thus a good generality to handle the data compression in various deep learning environments.

% \subsection{Fine-tune the Piece-wise Linear Mapping}
% \tao{what is the purpose of this evaluation (Fig.~\ref{plm})?
% Fig.~\ref{plm} will be combined with Fig.6 for a new section--Tuning the Piece-wise Linear Mapping, before evaluation.}

\subsection{Power Consumption}
\begin{figure}[t]
\begin{centering}
\includegraphics[width=1\columnwidth]{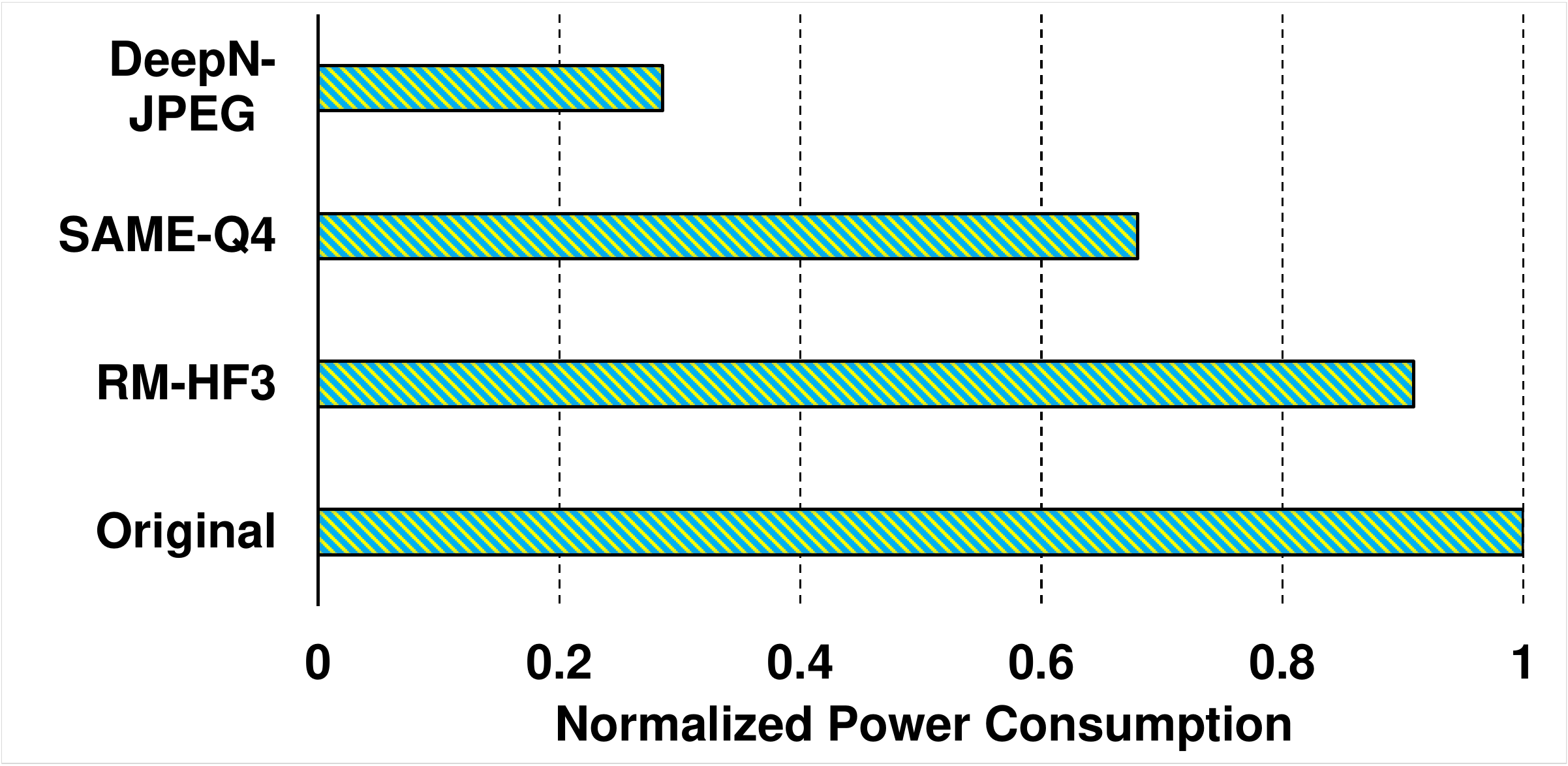}
\end{centering}
\vspace{-10pt}
\caption{Evaluation of power consumption for different methods.}
\label{power}
\vspace{-10pt}
\end{figure}

In resource-constraint terminal devices, the data offloading incurred power consumption can even outperform that of DNN computation in deep learning~\cite{kang2017neurosurgeon}. Date compression can reduce the associated cost. 
%now a major measurable component of system performance, especially for . 
%Therefore in this section we evaluate the power efficiency of proposed DeepN-JPEG. 
Following the same measurement in ~\cite{kang2017neurosurgeon}, Fig.~\ref{power} shows the results of power reduction breakdown. Our ``DeepN-JPEG" based data processing consumes only $30\%$ energy without accuracy reduction when compared with that of original dataset. Compared with ``RM-HF3'' (remove the top-3 high frequency components in quantization table) and ``SAME-Q4'' (the same quantization value--4 in quantization table), ``DeepN-JPEG" can still achieve $\sim2\times$ and $\sim3\times$ power reduction respectively, due to more efficient data compression. 
%thus a promising power efficiency.

\section{Conclusion}

The ever-increasing data transfer and storage overhead significantly challenges the energy efficiency and performance of large-scale DNNs. In this paper, we propose a DNN oriented image compression framework, namely ``DeepN-JPEG", to ease the storage and data communication overhead. Instead of the Human Vision System inspired JPEG compression, our solution effectively reduces the quantization error based on the frequency component analysis and rectified quantization table, and further increases the compress rate without any accuracy degradation. Our experimental results show that ``DeepN-JPEG'' achieves $\sim3.5\times$ compression rate improvement, and consumes only $30\%$ power of the conventional JPEG without classification accuracy degradation, thus a promising solution for data storage and communication for deep learning.

\bibliographystyle{IEEEtran}%{ACM-Reference-Format}
\bibliography{cites,new1} 

\end{document}